\documentclass[conference]{IEEEtran}

\usepackage{cite}
\usepackage{amsmath,amssymb,amsfonts}
\usepackage{algorithmic}
\usepackage{graphicx}
\usepackage{textcomp}
\usepackage{xcolor}
\usepackage{array,multirow}
\usepackage{tablefootnote}
\usepackage{threeparttable}
\usepackage{relsize}
\usepackage{enumitem}
\usepackage{amsmath}
\usepackage{url}
\usepackage{upgreek}
\usepackage[caption=false, font=footnotesize]{subfig}

\def\BibTeX{{\rm B\kern-.05em{\sc i\kern-.025em b}\kern-.08em
    T\kern-.1667em\lower.7ex\hbox{E}\kern-.125emX}}
    
\makeatletter
\IEEEtriggercmd{\reset@font\normalfont\fontsize{7.9pt}{8.60pt}\selectfont}
\newcommand*\titleheader[1]{\gdef\@titleheader{#1}}
\AtBeginDocument{%
  \let\st@red@title\@title
  \def\@title{%
    \bgroup\normalfont\normalsize\centering\@titleheader\par\egroup
    \vskip1ex\st@red@title}
}
\makeatother
\IEEEtriggeratref{1}

\titleheader{\color{gray}\vspace{-30pt}Presented at the 13th IEEE Latin America Symposium on Circuits and System (LASCAS), 01-04/03 2022, Puerto Varas, CL.\\This is the authors’ version of the paper. The final published version is posted on IEEE Xplore.\\ DOI: \url{https://doi.org/10.1109/LASCAS53948.2022.9789055}}

\title{MAx-DNN: Multi-Level Arithmetic Approximation for Energy-Efficient DNN Hardware Accelerators}

\graphicspath{{./FIGS/}}

\makeatletter
\def\ps@IEEEtitlepagestyle{
  \def\@oddfoot{\mycopyrightnotice}
  \def\@evenfoot{}
}
\def\mycopyrightnotice{
  {\footnotesize
  \begin{minipage}{\textwidth}
  \centering\color{gray}%
  ~\copyright~2022 IEEE.  Personal use of this material is permitted.  Permission from IEEE must be obtained for all other uses, in any current or future media, including reprinting/republishing this material for advertising or promotional purposes, creating new collective works, for resale or redistribution\\to servers or lists, or reuse of any copyrighted component of this work in other works.
  \end{minipage}
  }
}

\begin{document}

\author{\IEEEauthorblockN{Vasileios Leon\IEEEauthorrefmark{1}, 
    Georgios Makris\IEEEauthorrefmark{1}, 
    Sotirios Xydis\IEEEauthorrefmark{2}, 
    Kiamal Pekmestzi\IEEEauthorrefmark{1},
    Dimitrios Soudris\IEEEauthorrefmark{1}}\\[-11pt]
    \IEEEauthorblockA{\IEEEauthorrefmark{1}\emph{National Technical University of Athens, School of Electrical and Computer Engineering, 15780 Athens, Greece}}\\[-11pt]
    \IEEEauthorblockA{\IEEEauthorrefmark{2}\emph{Harokopio University of Athens, Department of Informatics and Telematics, 17778 Athens, Greece}}}

\maketitle

\begin{abstract}
Nowadays, the rapid growth of Deep Neural Network (DNN) architectures has established them as the defacto approach for providing advanced Machine Learning tasks with excellent accuracy. Targeting low-power DNN computing, this paper examines the interplay of fine-grained error resilience of DNN workloads in collaboration with  hardware approximation techniques, to achieve higher levels of energy efficiency. Utilizing the state-of-the-art ROUP approximate multipliers, we systematically explore their fine-grained distribution across the network according to our layer-, filter-, and kernel-level approaches, and examine their impact on accuracy and energy. We use the ResNet-8 model on the CIFAR-10 dataset to evaluate our approximations. The proposed solution delivers up to 54\% energy gains in exchange for up to 4\% accuracy loss, compared to the baseline quantized model, while it provides 2$\mathbf{\times}$ energy gains with better accuracy versus the state-of-the-art DNN approximations.
\end{abstract}


\section{Introduction}
The proliferation of demanding workloads in the 
Digital Signal Processing (DSP)
and Artificial Intelligence (AI) domains
marks
a new era in the design of integrated circuits and embedded systems.
Towards high-performance computing
within a limited power envelope,  
the research community is examining
alternative design strategies.
In this direction,
\emph{Approximate Computing} is an emerging design paradigm \cite{ac},
which exploits the error resilience
to trade accuracy for gains in power, energy, area, and/or latency.
In the field of inexact hardware,
approximation techniques are applied at different levels,
i.e., 
from circuits \cite{LeonDAC, auger, vleon, evolib} to accelerators \cite{leonfpl, alwann, approxcnn, axc,  approxqam}.

The intrinsic error resilience and increased computational demands 
of DNNs,
which are a state-of-the-art AI approach to tackle Computer Vision tasks (e.g.,
image classification, object detection, pose estimation),
constitutes them as promising candidates for approximate computing.
Typical
approximation techniques on DNNs
involve 
inexact arithmetic operators \cite{alwann},
low-bit numerical formats \cite{numeric}, 
weight quantization \cite{quantization}, 
and
neuron connection pruning \cite{pruning}.
Approximate hardware accelerators
for DNN inferencing 
\cite{alwann, approxcnn, axc}
have gained momentum 
due to their attractive performance-per-power ratio.
Moreover,
the ASIC/FPGA technology 
facilitates approximate computing
by 
allowing low-level optimizations
and custom datapath bit-widths,
contrary to the general-purpose GPU/CPU solutions.

The design of approximate DNN accelerators imposes several challenges.
As discussed in \cite{alwann},
one of the research goals is to avoid
retraining for reducing the accuracy loss.
The reason is twofold:
(i) proprietary datasets/models may not be available,
(ii) the increased training time
due the emulation of the hardware approximations
and 
the use of custom approximate arithmetic operators.
Another important challenge in approximate DNNs is the \textit{approximation localization}, i.e., where to insert approximations,
in order to maximize the energy efficiency
while keeping accuracy in acceptable levels. 

In this paper, 
we introduce the MAx-DNN  framework 
to address the aforementioned design challenges,
by examining approximation 
at different levels of the DNN architecture,
i.e., 
in
network's layers, 
layer's filters, 
or
filter's kernels.
Considering that the majority of the computations
in DNNs
are multiply-accumulate \cite{axc},
we focus on optimizing the multiplications.
MAx-DNN adopts
ALWANN \cite{alwann} for generating approximate DNN hardware accelerators without retraining,
and extends it with our approximation localization,
which is based on 
the ROUP multiplication library \cite{LeonDAC}.
Our design space is larger
compared to that of the original ALWANN framework,
which introduces the same inexact multipliers in each convolutional layer (multipliers may differ among layers).
Interestingly, 
the proposed approximation approaches
assign different approximate ROUP multipliers
either
in each convolutional layer, filter, or kernel.
To evaluate our approximations,
we employ the quantized ResNet-8 network,
trained
on the CIFAR-10 dataset,
and use
standard-cell ASIC technology 
with the TSMC 45-nm library.
At first, we examine the sensitivity of the DNN layers to approximations
in both standalone and combined fashion,
and then,
we perform an extensive design space exploration and 
accuracy--energy Pareto analysis
to extract the most prominent configurations of our approaches.
The results show that, 
compared to the quantized model,
our designs deliver
up to 54\% energy gains in exchange for an accuracy loss of 4\%,
while providing 2$\times$ energy efficiency versus the default ALWANN approximations.
The paper contribution lies in 
proposing an improved fine-grained method 
to approximate the DNN multiplications,
and
quantifying the results from various approximation approaches and configurations.

The remainder of the paper is organized as follows. 
Section II introduces the ALWANN framework \cite{alwann}.
Section III defines our approximation space.
Section IV reports the experimental evaluation. 
Finally, Section V draws the conclusions.

\section{The ALWANN Approximation Framework}
The ALWANN framework \cite{alwann} takes the following inputs:
(i) the trained (frozen) network model in protobuf format,
(ii) a library of approximate multipliers,
and
(iii) architecture constraints for the hardware accelerator (e.g., pipelined or power-gated mode, number of approximate units).
It assumes 
accurate addition and approximate multiplication, 
as well as one approximation type per convolutional layer.
Moreover,
to improve the accuracy without re-training, 
the network weights are tuned/updated
according to the multipliers' properties. 
The approximate networks,
labeled as AxNNs,
are modified versions of the initial model and
satisfy the user constraints for the architecture of the accelerator.

To enable ALWANN,
TensorFlow is extended 
to support approximate quantized layers 
by creating a new operator, 
which replaces 
the conventional \texttt{\small QuantizedConv2D} layers 
with \texttt{\small AxConv2D} layers.
This operator allows 
to specify which approximate multipliers to employ via the  \emph{AxMult(str)} parameter
(a C/C++ model of the multipliers is necessary),
and optionally use the weight tuning feature via \emph{AxTune(bool)}.
The frozen model is processed by the TensorFlow \emph{transform\_graph} tool,
which inserts the \texttt{\small AxConv2D} layers,
and then, 
the Pareto-optimal AxNNs are extracted by the NSGA-II algorithm.

\begin{figure}
    \centering
\includegraphics[width=1.0\columnwidth]{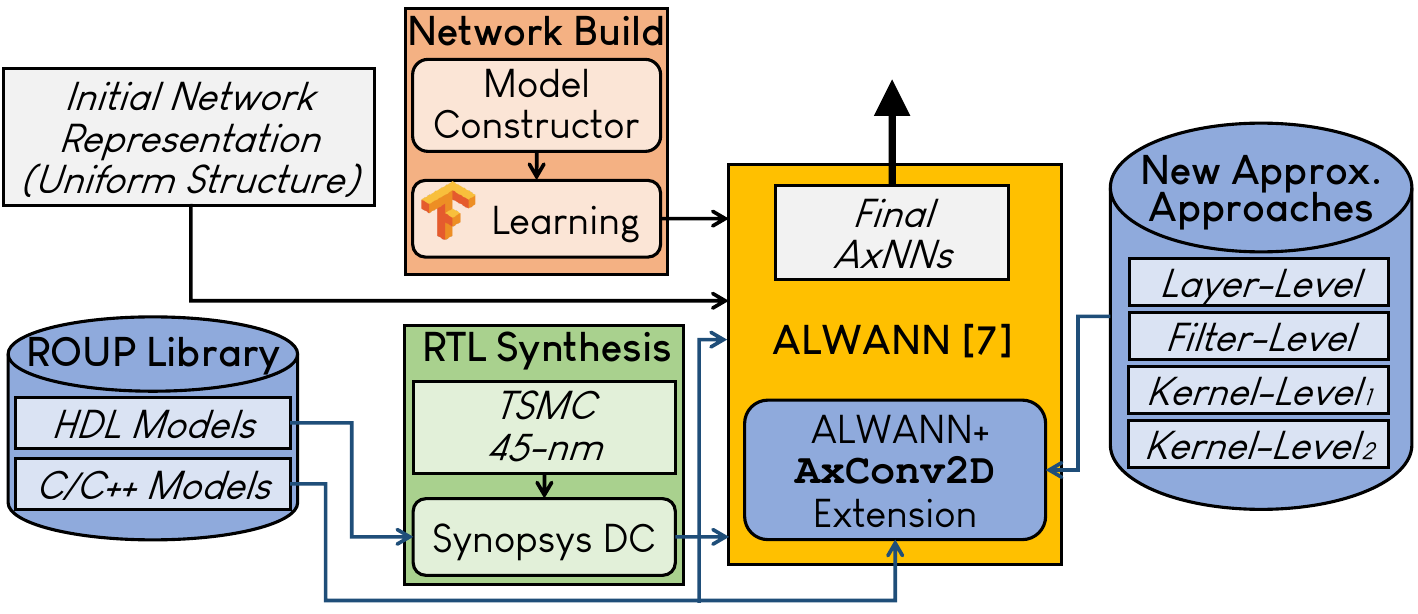}\\[-5pt]
    \caption{The MAx-DNN toolflow \& architecture, extension of ALWANN \cite{alwann}.}
    \label{alw_new}
    \vspace{-8pt}
\end{figure}

In this paper,
we equip ALWANN with 
new approximation approaches
regarding the distribution of the approximate multiplications across the network.
The toolflow of the extended framework version is illustrated in Fig. \ref{alw_new}.
In comparison with ALWANN,
we modify the \texttt{\small AxConv2D} TensorFlow operator to support our approximation approaches at three distinct levels (layer, filter, kernel)
and also employ the state-of-the-art ROUP library of approximate multipliers \cite{LeonDAC}.

\begin{figure*}[!t]
\centering
\hspace*{-10pt}
\subfloat[LLAM\label{t1}]{
\includegraphics[width=0.51\columnwidth]{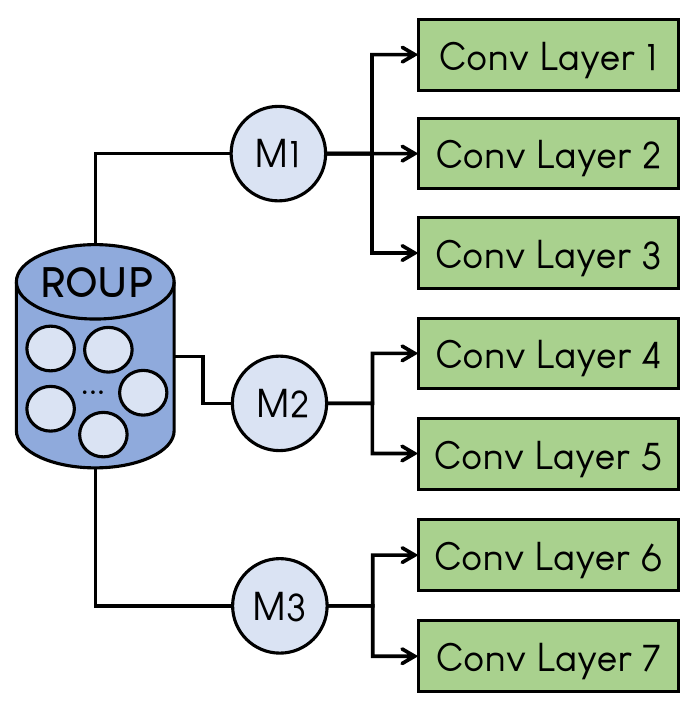}
}
\hspace*{-8pt}
\subfloat[FLAM\label{t2}]{
\includegraphics[width=0.51\columnwidth]{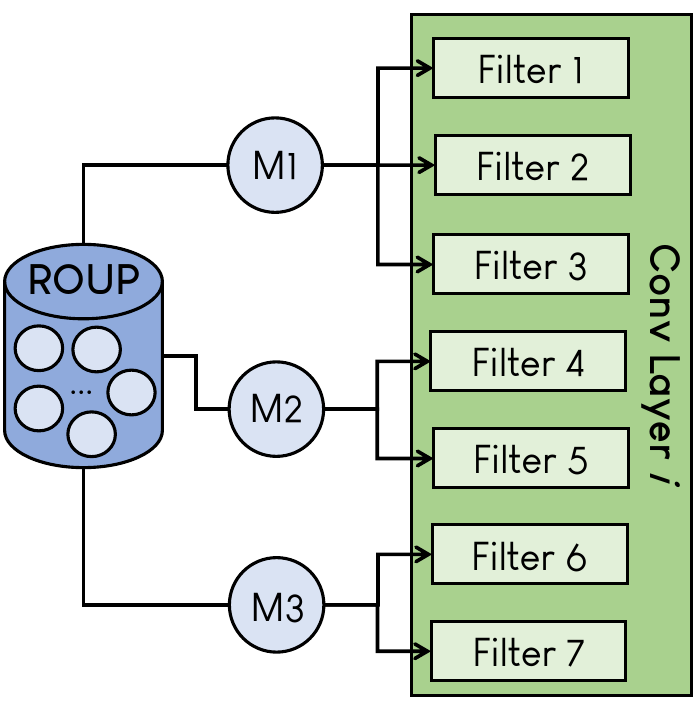}
}
\hspace*{-7pt}
\subfloat[KLAM\label{t3}]{
\includegraphics[width=0.51\columnwidth]{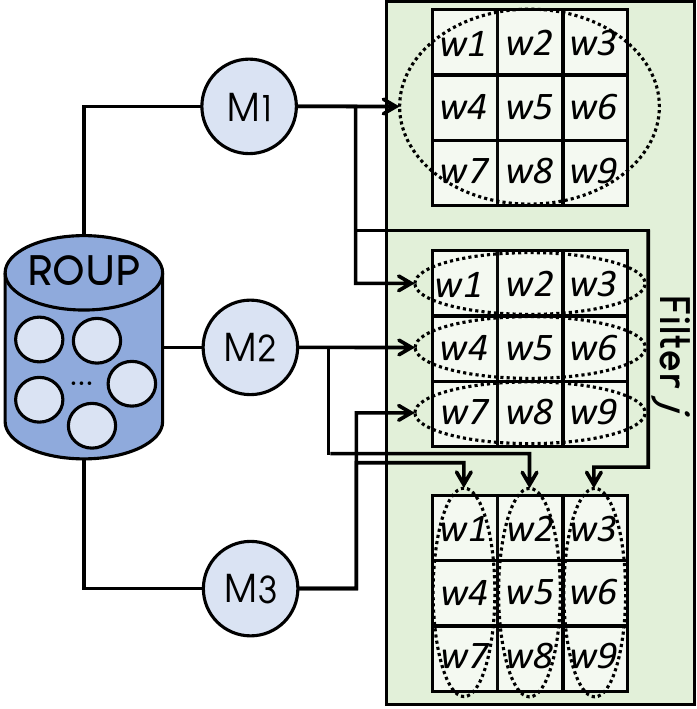}
}
\hspace*{-5pt}
\subfloat[KLMS\label{t4}]{
\includegraphics[width=0.51\columnwidth]{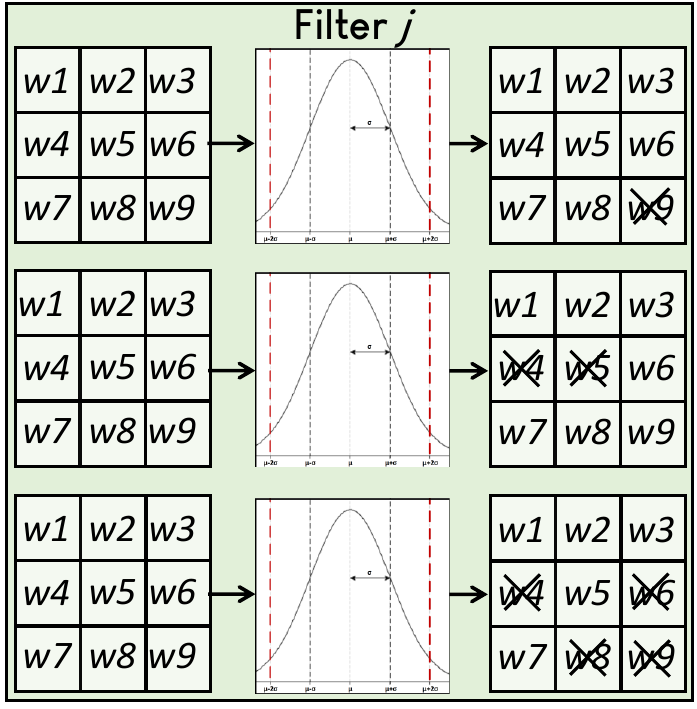}
}
\caption{The proposed non-uniform approximation approaches at different levels:
(a) layer-level,
(b) filter-level
\&
(c) kernel-level
approximate multiplication,
and 
(d) kernel-level multiplication skip.}
\label{tech}
\vspace{-10pt}
\end{figure*}

\section{Defining the Approximation Space}
In this section,
we discuss the proposed approximation space, explaining how we distribute the approximate multipliers
in each
layer, filter, or kernel of the DNN.
For our analysis,
we assume the typical DNN architecture:
each convolutional layer
includes $N$ filters,
with each one consisting of $M$ convolution kernels
to process the $M$ input channels 
and output a feature map ($N$ output feature maps are generated in total).
The approximations are not applied uniformly across the examined level,
i.e., each layer/filter/kernel is assigned a different ROUP multiplier.
Finally,
we present the ROUP multiplier family \cite{LeonDAC} and the model for estimating energy.

\subsection{LLAM: Layer-Level  Approximate Multiplication}
The first approach,
illustrated in Fig. \ref{t1}, 
aims to insert approximations with different strength in the network.
Specifically,
we perform all the multiplications of each convolutional layer
with a ROUP multiplier.
This approach is the conventional one,
where the approximations are distributed across the network
by assigning an approximate multiplier,
either the same (uniform distribution)
or different (non-uniform distribution),
to each convolutional layer.

\subsection{FLAM: Filter-Level Approximate Multiplication}
Similarly to LLAM,
our second approach,
illustrated in Fig. \ref{t2}, 
creates filters with different approximation for each convolutional layer of the DNN.
Namely,
we use different ROUP multipliers
in each filter of the layer,
contrary to the first approach,
where all the filters of a layer have the same ROUP multiplier.
This approach is implemented by creating groups of filters 
and assigning them the ROUP multipliers. 

\subsection{KLAM: Kernel-Level Approximate Multiplication}
In the third approximation approach,
we proceed deeper in the DNN architecture,
and approximate the multiplications  separately
for each convolutional kernel.
As a result, 
the convolutions of each filter  
are performed with different ROUP multipliers.
Fig. \ref{t3} illustrates the three proposed flavors of this approach:
the channel flavor,
where all
the multiplications corresponding to the kernel are performed with the same multiplier,
and the 
row/column flavor,
where 
different multipliers are employed for each row/column of the kernel.

\subsection{KLMS: Kernel-Level Multiplication Skip}
Our last approach,
illustrated in Fig. \ref{t4},
skips some of the multiplications of each convolution
based on the weight values.
More specifically,
we 
calculate the mean value of all the kernel weights in each layer,
and perform only the multiplications corresponding to the weights that are near it.
Assuming 
a mean $\mu$ and a standard deviation $\sigma$,
we select to perform only the multiplications with the weights belonging either in the interval
$[\mu-\sigma, \mu+\sigma]$ 
or
$[\mu-2\sigma, \mu+2\sigma]$. 

\subsection{The ROUP Approximate Multipliers}
The ROUP family of approximate multipliers \cite{LeonDAC}
is based on the radix-4 encoding,
and applies
two orthogonal approximation techniques,
i.e., 
\emph{Asymmetric Rounding} and \emph{Perforation}.
In this hybrid approximation,
\emph{Perforation} omits the generation of least-significant partial products,
and 
\emph{Asymmetric Rounding}
rounds each remaining partial product
to a different bit-width,
depending on its significance in the partial product matrix.

The ROUP multiplication of two $N$-bit 2’s-complement numbers
is performed by accumulating the non-perforated, rounded partial products,
i.e., 
$\tilde{P}_j$,
which are generated based on the radix-4 encoding \cite{LeonDAC}:
\begin{equation}
\text{ROUP}(A,B) = 
\mathlarger{\sum}_{\footnotesize{\substack{j=P}}}^{\footnotesize{\substack{N/2-1}}} 
\tilde{P}_j 4^{j} =
\mathlarger{\sum}_{\footnotesize{\substack{j=P}}}^{\footnotesize{\substack{N/2-1}}} 
A^r_j \! \cdot \!
b_{j}^{\text{\scriptsize R4*}} \! \cdot \!
4^{j}   
\end{equation}
The parameter $P$ denotes the number of successive least-significant partial products 
that are perforated.
For each remaining partial product $\tilde{P}_j$,
the input operand $A$
is rounded to its $r$ least-significant bit 
($r$ differs per product)
as follows:
\begin{equation}
A^r_j = \langle a_{N-1} a_{N-2} \cdots a_r\rangle_{2\text{'s}} + a_{r-1}  
\end{equation}
Bit-level manipulations in $A^r_j$
facilitate the process of rounding,  
as the above addition is absorbed in the calculations of the radix-4 correction terms,
and the carry propagation is avoided.
Finally,
$b_{j}^{\text{\scriptsize R4*}}$
is the conventional radix-4 logic function of the encoding signals of $B$,
with an extra XOR gate 
involving $a_{r-1}$,
which is 
imposed by the 
optimizations to avoid the addition.

The ROUP multipliers provide a large design space to explore various error--energy trade-offs,
as the approximations are tuned by two independent parameters.
In the comparative evaluation of \cite{LeonDAC},
the ROUP family formed the error--energy Pareto front with increased resolution,
outperforming several state-of-the-art multipliers.
In comparison with the multipliers from the EvoApprox8b library \cite{evolib},
which is used in ALWANN, 
ROUP provides more approximation configurations, 
especially for larger bit-widths.
As a result,
its error scaling is more dense, 
while in terms of energy,
several ROUP configurations outperform the EvoApprox8b multipliers by up to 15\%.

\subsection{Energy Model for Approximate DNNs}
The energy consumption of the 
approximate DNN accelerators
is estimated based on the implementation of the approximate multipliers.
Targeting standard-cell ASIC technology,
the multipliers are implemented 
on the TSMC 45-nm library
using industrial-strength tools, 
i.e.,
Synopsys Design Compiler for synthesis
and
Synopsys PrimeTime for measuring power.

Similarly to ALWANN \cite{alwann},
we calculate
the energy of each layer based on the number of multiplications and the energy per multiplier, 
i.e., $\#mult \times avg\_mult\_energy$.
To estimate the energy of the entire DNN,
we accumulate the individual layer energies.

\section{Experimental Evaluation}

To evaluate our approach, 
we employ the ResNet-8 DNN and the open-source ALWANN  framework \cite{alwann}, which is based on TensorFlow 1.14. 
ResNet-8 is trained with quantization on the CIFAR-10 dataset, 
achieving 83\% classification accuracy.

\begin{figure}[!b]
\centering
\hspace*{-9pt}
\subfloat[\label{y1}]{\includegraphics[width=0.55\columnwidth]{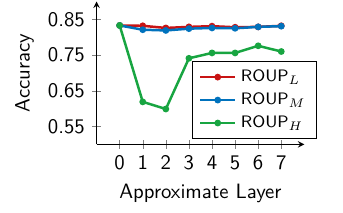}
}\hspace*{-16pt}
\subfloat[\label{y2}]{\includegraphics[width=0.55\columnwidth]{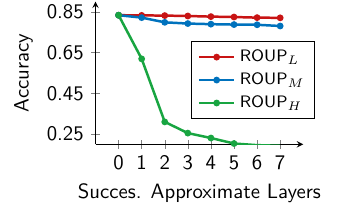}}
\caption{The scaling of ResNet-8 accuracy w.r.t. the layers approximated: 
    (a) one layer per occasion (e.g., only the 5th), 
    and (b) successive layers (e.g., layers 1 to 5). ``Layer=0'' denotes the baseline quantized model.}
\label{yy}
\end{figure}

\textbf{Layer-wise study of error resilience:} At first, we examine the error sensitivity of the convolutional layers,
in an effort to 
understand which layers are offered for approximations.
For this experiment,
we pick three ROUP multipliers with different approximation strength,
i.e., ``low'', ``medium'', and ``high'',
labeled as ROUP$_L$, ROUP$_M$, and ROUP$_H$, respectively. 
Fig. \ref{y1} illustrates the accuracy scaling
when using these multipliers 
only in the $m$-th convolutional layer
($m$ = 1, 2,$\scriptstyle \,\, \dots$7).
Regardless of the approximation strength,
it is shown that 
approximating one of the first layers
results in remarkable accuracy loss
($m$ = 0 shows the baseline model with 83\% accuracy).
This is more evident in the ROUP$_H$ configurations,
where significant computation errors are inserted.
In this case, 
when approximating one of the layers 4--7, 
the accuracy loss is decreased
and stabilized around 8\%. 
Fig. \ref{y2},
depicts the accuracy scaling
when approximating the first $m$ layers.
As expected,
the accuracy loss is increased
with respect to the number of approximate layers,
however,
we again notice 
the error resilience of the last layers.
Specifically,
the accuracy loss is slowing down 
when extending the approximation after the 4-th layer.
Another important outcome from this exploration
is the negligible accuracy loss of the ROUP$_L$ and ROUP$_M$ configurations,
regardless of which and how many layers are approximated,  
as these multipliers 
provide significant energy gains compared to their accurate design,
i.e., around 10\% and 20\%, respectively.

\begin{figure}[!t]
    \centering
    \vspace{-8pt}
  \hspace*{-10.5pt} 
\includegraphics[width=1.05\columnwidth]{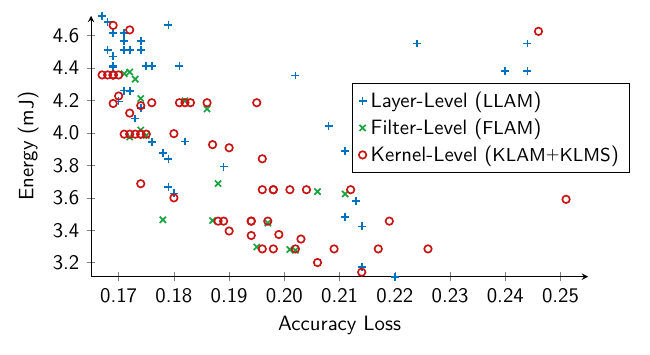}\\[-12pt]
    \caption{The energy consumption and accuracy loss of the approximate ResNet-8 hardware accelerators, implemented with the proposed approaches.}
    \label{pareto}
    \vspace{-10pt} 
\end{figure}

\textbf{Exploration of approximation space:} Subsequently,
we perform an extensive design space exploration on the proposed approaches,
involving various ROUP multipliers,
to extract their most prominent
configurations in terms of accuracy and energy.
For each approach,
several multiplication replacements and combinations are examined. 
In Fig. \ref{pareto}, 
we present
all the configurations that deliver at least 75\% classification accuracy, 
i.e., up to 8\% accuracy loss compared to the quantized model. 
As shown in the upper left segment of the plot, 
multiple configurations 
deliver negligible accuracy loss compared to the quantized model,
which ranges from 0.02\% to 1\%,
while 
improving the energy efficiency due to using the ROUP multipliers.
Therefore,
our approaches can satisfy near-zero accuracy loss
with more energy-efficient computing.
Regarding the efficiency of each approach,
the Pareto front is formed almost exclusively from the kernel- and filter-level configurations. 
For the same accuracy loss,
these configurations
provide better energy
than the conventional layer-level approach,
which has to sacrifice a large amount of accuracy,
i.e., more than 40\%,
to deliver this energy efficiency.

\textbf{Comparison to ALWANN framework:} Table \ref{tbb} compares the 
Pareto-front configurations
with ALWANN networks employing the EvoApprox8b multipliers \cite{evolib}.
The proposed designs deliver better accuracy,
as the average loss of the
EvoApprox8b configurations
is \raisebox{0.8pt}{$\scriptstyle\sim$}23\%, 
while
in terms of energy,
they provide 2$\times$ gains.
This comparison highlights the advantage 
of studying the approximations
at a lower DNN level, i.e., filter or kernel.
Namely, 
the fine-grained use of multipliers with different approximation strength
outperforms the conventional approximation
of all the layer multiplications.

\vspace{5pt}
\noindent\textbf{Lessons Learnt:} 
According to our multi-level design space exploration
and evaluation,
we experimentally prove that:
\begin{enumerate}[wide=0pt, label=(\roman*)]
\item the first convolutional layers are more sensitive in approximations,
i.e., less error resilient, 
than the final ones.
\item the approximation of multipliers based on a non-uniform fine-grained filter/kernel-level approach delivers better accuracy 
than 
the coarse-grained layer-level approximation.
\end{enumerate}

\section{Conclusion and Future Work}
In this work,
we presented the MAx-DNN framework to explore the efficiency of multi-level arithmetic approximation in DNN hardware accelerators.
MAx-DNN 
extends prior-art design approaches considering approximate multiplier heterogeneity,
not only to each DNN layer, 
but also to each filter and kernel.
The results show that our most prominent configurations
achieve up to 54\% energy reduction and 4\% accuracy loss,  
while providing 2$\times$ gains versus the straightforward approximation with EvoApprox8b multipliers.
Our future work includes 
evaluation on different DNN types, 
and study of the error propagation among the DNN nodes. 

\begin{table}[!t]
\vspace{-3pt}
\renewcommand{\arraystretch}{1.15}
\centering
\setlength{\tabcolsep}{0pt}
\caption{Comparison of Approximate ResNet-8 Hardware Accelerators}
\vspace{-5pt}
 \begin{threeparttable}
\begin{tabular*}{\columnwidth}{@{} @{\extracolsep{\fill}}*{1}{l}
@{} 
@{\extracolsep{\fill}}*{1}{l}
@{}
@{\extracolsep{\fill}}*{2}{c} @{}}
\hline
\textbf{Approximation} & \textbf{Approach/Config.} & \textbf{Energy Gain} & \textbf{Accuracy Loss\footnotemark} \\
\hline
\hline
\parbox[t]{7mm}{\multirow{8}{*}{\rotatebox[origin=c]{45}{Proposed -- ROUP \cite{LeonDAC}}}} & FLAM-3clas.\_2\_1\_1 & 49\% & 18\%\\
& FLAM-3clas.\_2\_2\_1 & 52\% & 20\%\\
& KLAM-chan.\_1\_0\_1  & 46\% & 17\%\\
& KLAM-chan.\_2\_0\_2  & 53\% & 21\%\\
& KLAM-chan.\_1\_1\_2  & 50\% & 19\%\\
& KLAM-chan.\_2\_1\_2  & 54\% & 21\%\\
& KLAM-row\_2\_1\_1      & 50\% & 19\%\\
& KLAM-row\_2\_1\_2      & 52\% & 20\%\\
\parbox[t]{7mm}{\multirow{5}{*}{\rotatebox[origin=c]{45}{Uniform -- Evo \cite{evolib}}}} & Evo\_mul8\_2AC   & 23\% & 20\%\\
& Evo\_mul8u\_2HH  & 23\% & 23\%\\
& Evo\_mul8u\_NGR  & 32\% & 23\%\\
& Evo\_mul8u\_ZFB & 39\% & 23\%\\
& Evo\_mul8u\_7C1  & 20\% & 24\%\\ 
\hline
\end{tabular*}
 \begin{tablenotes}
  \item[1]{\scriptsize It is reported compared to the full accurate model. The baseline quantized model,
  where we apply our approximations,
  already has an accuracy loss of 17\%.} 
 \end{tablenotes}
 \end{threeparttable}
\label{tbb}
\vspace{-10pt}
\end{table}


\bibliographystyle{IEEEtran}
\bibliography{REFERENCES.bib}

\begin{thebibliography}{10}
\providecommand{\url}[1]{#1}
\csname url@samestyle\endcsname
\providecommand{\newblock}{\relax}
\providecommand{\bibinfo}[2]{#2}
\providecommand{\BIBentrySTDinterwordspacing}{\spaceskip=0pt\relax}
\providecommand{\BIBentryALTinterwordstretchfactor}{4}
\providecommand{\BIBentryALTinterwordspacing}{\spaceskip=\fontdimen2\font plus
\BIBentryALTinterwordstretchfactor\fontdimen3\font minus \fontdimen4\font\relax}
\providecommand{\BIBforeignlanguage}[2]{{%
\expandafter\ifx\csname l@#1\endcsname\relax
\typeout{** WARNING: IEEEtran.bst: No hyphenation pattern has been}%
\typeout{** loaded for the language `#1'. Using the pattern for}%
\typeout{** the default language instead.}%
\else
\language=\csname l@#1\endcsname
\fi
#2}}
\providecommand{\BIBdecl}{\relax}
\BIBdecl

\bibitem{ac}
S.~Mittal, ``{A Survey of Techniques for Approximate Computing},'' \emph{ACM Computing Surveys}, vol.~48, no.~4, Mar. 2016.

\bibitem{LeonDAC}
V.~Leon \emph{et~al.}, ``{Cooperative Arithmetic-Aware Approximation Techniques for Energy-Efficient Multipliers},'' in \emph{Design Automation Conference (DAC)}, 2019, pp. 1--6.

\bibitem{auger}
D.~Hernandez-Araya \emph{et~al.}, ``{AUGER: A Tool for Generating Approximate Arithmetic Circuits},'' in \emph{IEEE Latin American Symposium on Circuits Systems (LASCAS)}, 2020, pp. 1--4.

\bibitem{vleon}
V.~Leon \emph{et~al.}, ``{Approximate Hybrid High Radix Encoding for Energy-Efficient Inexact Multipliers},'' \emph{IEEE Transactions on Very Large Scale Integration (VLSI) Systems}, vol.~26, no.~3, pp. 421--430, Mar. 2018.

\bibitem{evolib}
V.~Mrazek \emph{et~al.}, ``{EvoApprox8b: Library of Approximate Adders and Multipliers for Circuit Design and Benchmarking of Approximation Methods},'' in \emph{Design, Automation and Test in Europe Conference (DATE)}, 2017, pp. 258--261.

\bibitem{leonfpl}
V.~Leon \emph{et~al.}, ``{Exploiting the Potential of Approximate Arithmetic in DSP \& AI Hardware Accelerators},'' in \emph{Int’l. Conference on Field Programmable Logic and Applications (FPL)}, 2021, pp. 1--2.

\bibitem{alwann}
V.~Mrazek \emph{et~al.}, ``{ALWANN: Automatic Layer-Wise Approximation of Deep Neural Network Accelerators without Retraining},'' in \emph{Int’l. Conference on Computer-Aided Design (ICCAD)}, 2019, pp. 1--8.

\bibitem{approxcnn}
G.~Lentaris \emph{et~al.}, ``{Combining Arithmetic Approximation Techniques for Improved CNN Circuit Design},'' in \emph{Int’l. Conference on Electronics, Circuits and Systems (ICECS)}, 2020, pp. 1--4.

\bibitem{axc}
S.~Venkataramani \emph{et~al.}, ``{Efficient AI System Design With Cross-Layer Approximate Computing},'' \emph{Proceedings of the IEEE}, vol. 108, no.~12, pp. 2232--2250, Dec. 2020.

\bibitem{approxqam}
V.~Leon \emph{et~al.}, ``{ApproxQAM: High-Order QAM Demodulation Circuits with Approximate Arithmetic},'' in \emph{Int’l. Conference on Modern Circuits and Systems Technologies (MOCAST)}, 2021, pp. 1--5.

\bibitem{numeric}
U.~K\"{o}ster \emph{et~al.}, ``{Flexpoint: An Adaptive Numerical Format for Efficient Training of Deep Neural Networks},'' in \emph{Int’l. Conference on Neural Information Processing Systems (NIPS)}, 2017, pp. 1740--1750.

\bibitem{quantization}
J.~Wu \emph{et~al.}, ``{Quantized Convolutional Neural Networks for Mobile Devices},'' in \emph{IEEE Conference on Computer Vision and Pattern Recognition (CVPR)}, 2016, pp. 4820--4828.

\bibitem{pruning}
S.~Anwar \emph{et~al.}, ``{Structured Pruning of Deep Convolutional Neural Networks},'' \emph{ACM Journal on Emerging Technologies in Computing Systems (JETC)}, vol.~13, no.~3, Feb. 2017.

\end{thebibliography}

\end{document}